%% file: main.tex
\documentclass{article}

\usepackage{microtype}
\usepackage{graphicx}
\usepackage{booktabs} % for professional tables

\usepackage{hyperref}

\usepackage{amsmath}
\usepackage{multicol}
\usepackage{amsfonts}
\usepackage{dsfont}
\usepackage{tikz}
\usetikzlibrary{shapes.multipart}
\usepackage{xcolor}
\usepackage{caption}
\usepackage{soul}
\usepackage{subcaption}
\newcommand{\DKL}[2]{D_{\mathrm{KL}} \left( #1 \left|\right| #2 \right) }

\usepackage{graphicx}
\usepackage{wrapfig}
\graphicspath{ {./Assets/} }

\usepackage[accepted]{icml2021}

\icmltitlerunning{Goal-Conditioned Reinforcement Learning with Imagined Subgoals}

\begin{document}

\twocolumn[
\icmltitle{ Goal-Conditioned Reinforcement Learning with Imagined Subgoals}



\icmlsetsymbol{equal}{*}

\begin{icmlauthorlist}
\icmlauthor{Elliot Chane-Sane}{to}
\icmlauthor{Cordelia Schmid}{to}
\icmlauthor{Ivan Laptev}{to}
\end{icmlauthorlist}

\icmlaffiliation{to}{Inria, \'{E}cole normale suprieure, CNRS, PSL Research University, 75005 Paris, France}

\icmlcorrespondingauthor{Elliot Chane-Sane}{elliot.chane-sane@inria.fr}

\icmlkeywords{Reinforcement Learning, Robotics}

\vskip 0.3in
]



\printAffiliationsAndNotice{}  % leave blank if no need to mention equal contribution

\input{Sections/Introduction}
\input{Sections/RelatedWork}

\input{Sections/Method}

\input{Sections/Experiments}
\input{Sections/Conclusion}

\bibliography{biblio}
\bibliographystyle{icml2021}

\newpage

\appendix
\onecolumn
\input{Appendix/Derivation}
\input{Appendix/Implementation_details}
\input{Appendix/Environments}
\input{Appendix/Hyperparameters}

\input{Appendix/AdditionalResults}
%%%%%%%%%%%%%%%%%%%%%%%%%%%%%%%%%%%%%%%%%%%%%%%%%%%%%%%%%%%%%%%%%%%%%%%%%%%%%%%
%%%%%%%%%%%%%%%%%%%%%%%%%%%%%%%%%%%%%%%%%%%%%%%%%%%%%%%%%%%%%%%%%%%%%%%%%%%%%%%

\end{document}

%% file: Sections/Introduction.tex
\begin{abstract}

Goal-conditioned reinforcement learning endows an agent with a large variety of skills, but it often struggles to solve tasks that require more temporally extended reasoning.
 In this work, we propose to incorporate imagined subgoals into policy learning to facilitate learning of complex tasks. Imagined subgoals are predicted by a separate high-level policy, which is trained simultaneously with the policy and its critic. This high-level policy predicts intermediate states halfway to the goal using the value function as a reachability metric. We don’t require the policy to reach these subgoals explicitly. Instead, we use them to define a prior policy, and incorporate this prior into a KL-constrained policy iteration scheme to speed up and regularize learning.  Imagined subgoals are used during policy learning, but not during test time, where we only apply the learned policy.  We evaluate our approach on complex robotic navigation and manipulation tasks and show that it outperforms existing methods by a large margin.
\end{abstract}

\section{Introduction}

\begin{figure}[t]
\centering
%\begin{subfigure}{0.5\columnwidth}
\includegraphics[width=1.0\linewidth]{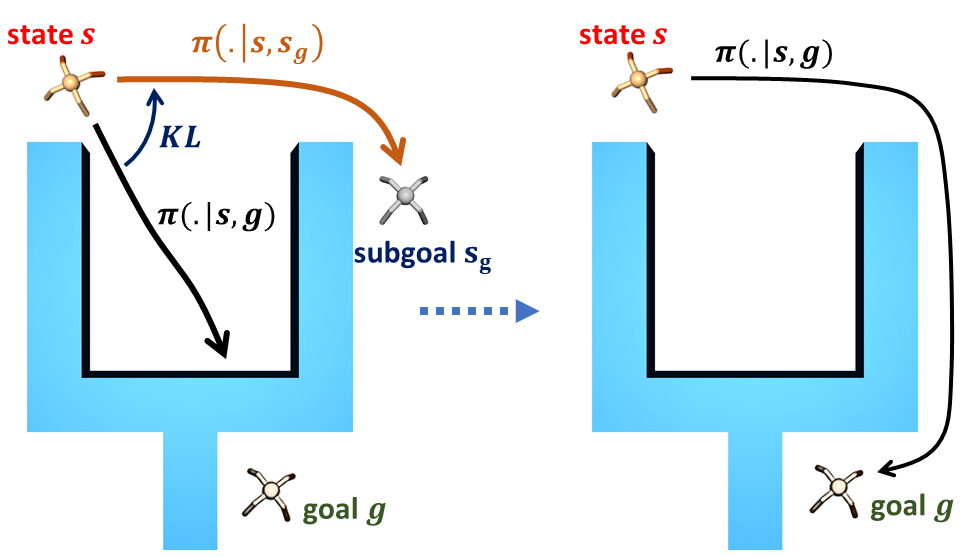}
\caption{
Illustration of the KL-regularized policy learning using imagined subgoals. 
(Left): The policy fails to reach a distant goal, yet it can reach a closer subgoal. Our approach automatically generates imagined subgoals for a task and uses such subgoals to direct the policy search during training.  
(Right): At test time, the resulting flat policy can reach arbitrarily distant goals \textit{without} relying on subgoals. 
\vspace{-.3cm}}
\label{fig:opener}
\end{figure}

An intelligent agent aims at solving tasks of varying horizons in an environment. It must be able to identify how the different tasks intertwine with each other  and leverage behaviors learned by solving simpler tasks to efficiently master more complex tasks.
For instance, once a legged robot has learned how to walk in every direction in a simulated maze environment, it could use these behaviors to efficiently learn to navigate in this environment.
Goal-conditioned reinforcement learning (RL) \cite{Kaelbling1993LearningTA, pmlr-v37-schaul15, andrychowicz2017hindsight} 
defines each task by the desired goal and, in principle, could learn a wide range of skills. 
In practice, however, reinforcement learning struggles to perform temporally extended reasoning.

Hierarchical methods have proven effective for  learning temporal abstraction in long-horizon problems \cite{NIPS1992_d14220ee, wiering1997hq,  sutton1999between,dietterich2000hierarchical}.
In the goal-conditioned setting, a high-level controller typically finds an appropriate sequence of subgoals that can be more easily followed by a low-level policy~%can more easily reach one after another 
\cite{nachum2018data, Gupta2019RelayPL}. 
When chosen appropriately, these subgoals effectively decompose a complex task into easier  tasks.
However, hierarchical methods are often unstable to train \cite{nachum2018data} and rely on appropriate temporal design choices.

In this work, we propose to use subgoals to improve the goal-conditioned policy.
Instead of reaching the subgoals explicitly, our method builds on the following intuition. If the current policy can readily reach a subgoal, it  could provide  guidance for reaching more distant goals,  as illustrated in Figure \ref{fig:opener}. 
We apply this idea to all possible state and goal pairs of the environment. 
%We apply this idea in an iterative fashion for increasingly distant goals. 
This self-supervised  approach progressively extends the horizon of the agent throughout training. At the end of training, the resulting flat policy does not require access to subgoals and can reach 
%any arbitrarily 
distant goals in its environment.

\begin{figure*}[ht]  
\centering
\begin{tikzpicture}[every text node part/.style={align=center}]
    % Policy
    \draw[very thick, rounded corners] (8,3.5) rectangle (11,5.0) node[pos=0.5, font=\bfseries]{Policy \\ $\pi$};
    \draw[very thick, rounded corners] (8.1,3.6) rectangle (10.9,4.9);
    % Prior policy
    %\fill[black!10, rounded corners, very thick, dotted] (2.75,0.75) rectangle (11.25,3.0);
    \draw[black, rounded corners, very thick, dashed] (2.75,0.75) rectangle (11.25,3.0);
    \node[black, font=\bfseries] at (10.0,0.5) {Prior policy};
    % High-level policy
    \draw[very thick, rounded corners] (3,1) rectangle (6,2.5) node[pos=0.5,black, font=\bfseries]{High-level policy \\ $\pi^H$};
    % Bootstrapped policy
    \draw[very thick, rounded corners] (8,1.25) rectangle (11,2.75) node[pos=0.5, font=\boldmath]{$\pi(.|s, s_g)$};
    \draw[very thick, rounded corners] (8.1,1.35) rectangle (10.9,2.65);
    %\draw[black, very thick, dotted, rounded corners] (8,1.25) rectangle (11,2.75);

    % State paths
    \draw[black!30!red, very thick, ->,font=\boldmath] (3.0,4.5) --  (8.0,4.5) node [at start, anchor=east] {state $s$};
    \draw[black!30!red, very thick, ->] (4.0,4.5) --  (4.0,2.5);
    \draw[black!30!red, very thick] (7.0,4.5) --  (7.0,2.25);
    \draw[black!30!red, very thick, ->] (7.0,2.25) --  (8.0,2.25);
    % Goal paths
    \draw[black!60!green, very thick, ->,font=\boldmath] (3.0,4.0) --  (8.0,4.0) node [at start, anchor=east] {goal $g$};
    \draw[black!60!green, very thick, ->] (5.0,4.0) --  (5.0,2.5);
    % Subgoal path
    \draw[blue, very thick, ->] (6.0,1.75) --  (8.0,1.75) node [midway, anchor=north,font=\boldmath] {subgoals $s_g$};
    % Bootstrapping path
    \draw[black, very thick, ->] (9.5,3.5) --  (9.5,2.75);
    % Policy paths
    \draw[very thick, ->] (11.0,4.25) --  (12.0,4.25) node [at end, anchor=west,font=\boldmath] {$\pi(.|s, g)$};
    \draw[very thick, ->] (11.0,2.0) --  (11.75,2.0) node [at end, anchor=west,font=\boldmath] {$\pi^{prior}(.|s, g)$};
    % KL regularization path
    \draw[black!60!blue, very thick, ->] (12.5,2.5) --  (12.5,3.75) node [font=\boldmath, midway, anchor=west] {$D_{\mathrm{KL}}$};
\end{tikzpicture}
\vspace{-.2cm}
\caption{
    Overview over our Reinforcement learning with Imagined Subgoals (RIS) approach. During policy training, the policy $\pi$ is constrained to stay close to the prior policy $\pi^{prior}$ through KL-regularization. 
    We define the prior policy $\pi^{prior}$ as the distribution of actions required to reach intermediate subgoals $s_g$ of the task.
    Given the initial state $s$ and the goal state $g$, the subgoals are generated by the high-level policy $\pi^H$. % that is learned jointly with $\pi$. 
    %To incorporate subgoals into the policy learning, we define our prior policy as the distribution over the actions the policy would have taken to reach intermediate subgoals predicted by the high-level policy.
    Note that the high-level policy and subgoals are only used during training of the target policy $\pi$. At test time we directly use $\pi$ to generate appropriate actions.
    %When interacting with the environment at test time, the learned policy directly maps states and goals to distributions over actions, \textit{without} using the high-level policy.
}
\vspace{-.2cm}
\label{fig:PriorPolicy}  
\end{figure*}
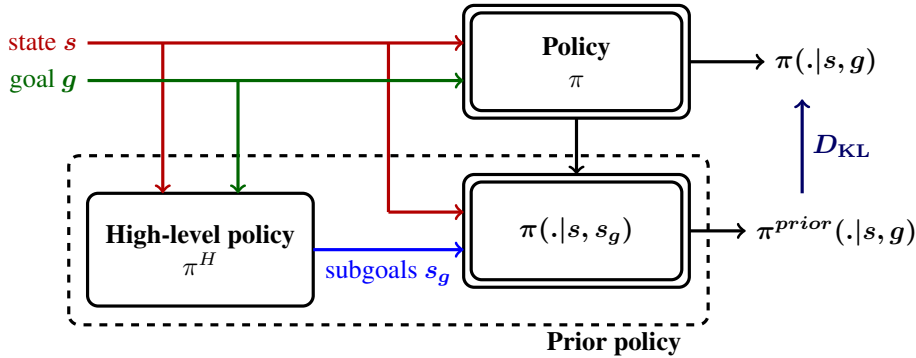 

Our method does not require a full sequence of subgoals, but predicts subgoals that are halfway to the goal, such that reaching them corresponds to a lower level of temporal abstraction than reaching the goal. 
To handle this subgoal prediction problem, we simultaneously train a separate high-level policy operating in the state space and use the value function of the goal-conditioned policy as a relative measure of distance between states \cite{eysenbach2019search, nasiriany2019planning}.

To incorporate subgoals into policy learning, we introduce a prior policy defined as the distribution over actions required to reach intermediate subgoals (Figure \ref{fig:PriorPolicy}).
When using appropriate subgoals that are easier to reach, this prior policy corresponds to an initial guess to reach the goal.
Accordingly, we leverage a policy iteration with an additional Kullback-Leibler (KL) constraint scheme towards this prior policy.
This adequately encourages the policy to adapt the simpler behaviors associated with reaching these subgoals to master more complex goal-reaching tasks in the environment.
Because these subgoals are \textit{not} actively pursued when interacting with the actual environment but only used to accelerate policy learning, we refer to them as \textit{imagined subgoals}. Imagined subgoals and intermediate subgoals have equal meaning in this paper.

Our method, Reinforcement learning with Imagined Subgoals (RIS), builds upon off-policy actor-critic approaches for continuous control and additionally learns a high-level policy that updates its predictions according to the current goal-reaching capabilities of the policy.
Our approach is general, solves temporally extended goal-reaching tasks with sparse rewards, and self-supervises its training by choosing appropriate imagined subgoals to accelerate its learning.

In summary, our contributions are threefold: 
(1) we propose a method for predicting subgoals which decomposes a goal-conditioned task into easier subtasks;
(2) we incorporate these subgoals into policy learning through a KL-regularized policy iteration scheme with a specific choice of prior policy;
and (3) we show that our approach greatly accelerates policy learning on a set of simulated robotics environments that involve motor control and temporally extended reasoning.\footnote{Code is available on the project webpage  \url{https://www.di.ens.fr/willow/research/ris/}.} 

%% file: Sections/RelatedWork.tex
\section{Related Work}
\label{section:related_work}

Goal-conditioned reinforcement learning has been addressed by a number of methods~\cite{Kaelbling1993LearningTA, pmlr-v37-schaul15, andrychowicz2017hindsight, Veeriah2018ManyGoalsRL, pong2018temporal, nair2018visual, zhao2019maximum, pitis2020maximum, eysenbach2020clearning}.
Given the current state and the goal, the resulting policies predict action sequences that lead towards the desired goal.
Hindsight experience replay (HER) \cite{Kaelbling1993LearningTA, andrychowicz2017hindsight} is often used to improve the robustness and sample efficiency of goal-reaching policies.
While in theory such policies can address any goal-reaching task, they often fail to solve temporally extended problems in practice~\cite{levy2017learning, nachum2018data}.

Long-horizon tasks can be addressed by hierarchical reinforcement learning~\cite{NIPS1992_d14220ee,wiering1997hq,dietterich2000hierarchical,levy2017learning, vezhnevets2017feudal}. Such methods often design  high-level policies that operate at a coarser time scale and control execution of low-level policies.
To address goal-conditioned settings, high-level policies can be learned to iteratively predict a sequence of intermediate subgoals. Such subgoals can then be used as targets for low-level policies~\cite{nachum2018data, Gupta2019RelayPL, nair2019hierarchical}.
As an alternative to the iterative planning, other methods generate sequences of subgoals with a divide-and-conquer approach~\cite{jurgenson2020sub, parascandolo2020divide,pertsch2020long}.
While hierarchical RL methods can better address long-horizon tasks, the joint learning of high-level and low-level policies may imply instabilities~\cite{nachum2018data}.
Similar to previous hierarchical RL methods, we use subgoals to decompose long-horizon tasks into simpler problems. Our subgoals, however, are only used during policy learning to guide and accelerate the search of the non-hierarchical policy.

Several recent RL methods use the value function of goal-reaching policies to measure distances between states and to plan sequences of subgoals \cite{nasiriany2019planning, eysenbach2019search, zhang2020world}.
In particular, LEAP~\cite{nasiriany2019planning} uses the value function of TDM policies \cite{pong2018temporal} and 
optimizes sequences of appropriate subgoals at test time. 
Similar to previous methods, we use the value function as a distance measure between states. Our method, however, avoids expensive test-time optimization of subgolas. We also experimentally compare our method with LEAP and demonstrate improvements. 
Moreover, we show that our approach can benefit from recent advances in learning representations for reinforcement learning from pixels on a vision-based robotic manipulation task \cite{kostrikov2020image, laskin2020reinforcement}.

Many approaches cast reinforcement learning as a probabilistic inference problem where the optimal policy should match a probability distribution in a graphical model defined by the reward and the environment dynamics \cite{toussaint2009robot, kappen2012optimal, levine2018reinforcement}.
Several methods optimize an objective incorporating the divergence between the target policy and a prior policy.
The prior policy can be fixed \cite{haarnoja2017reinforcement, abdolmaleki2018maximum, haarnoja2018soft, pertsch2020accelerating} 
or learned jointly with the policy \cite{teh2017distral, galashov2019information, tirumala2019exploiting}.
While previous work imposes explicit priors e.g., in the multi-task and transfer learning settings \cite{teh2017distral, galashov2019information, tirumala2019exploiting}, we constrain our policy search by the prior distribution implicitly defined by subgoals produced by the high-level policy. 

Behavior priors have often been used to avoid value overestimation for out-of-distribution actions in offline reinforcement learning \cite{fujimoto2018addressing, wu2019behavior,  kumar2019stabilizing, siegel2020keep, nair2020accelerating, wang2020critic}.
Similar to these methods, we constrain our high-level policy to avoid predicting subgoals outside of the valid state distribution.

% About Guided policy search
Finally, our work shares similarities with guided policy search methods  ~\cite{levine2013guided, levine2014learning, levine2016end} which alternate between generating expert trajectories using trajectory optimization and improving the learned policy.
In contrast, our policy search is guided by subgoals produced by a high-level policy.

%% file: Sections/Method.tex
\section{Method}

Our method builds on the following key observation. If an action $a$ is well-suited for approaching an intermediate subgoal $s_g$ from state $s$, it should also be a good choice for approaching the final goal $g$ of the same task. We assume that reaching subgoals $s_g$ is simpler than reaching more distant goals $g$. Hence, we adopt a self-supervised strategy and use the subgoal-reaching policy $\pi(\cdot|s,s_g)$ as a guidance when learning the goal-reaching policy $\pi(\cdot|s,g)$. We denote our approach as Reinforcement learning with Imagined Subgoals (RIS)
and present its overview in  Figures~\ref{fig:opener} and~\ref{fig:PriorPolicy}.
%The overview of RIS is presented in Figures~\ref{fig:opener},\ref{fig:PriorPolicy}.

To implement the idea of RIS, we first introduce a high-level policy $\pi^H$ predicting imagined subgoals $s_g$, as described in Section~\ref{subsec:hp}. Section~\ref{subsec:pi} presents the regularized learning of the target policy $\pi$ using subgoals. The joint training of $\pi$ and $\pi^H$ is summarized in Section~\ref{subsec:algo}. Before describing our technical contributions, we present the actor-critic paradigm used by RIS in Section~\ref{subsec:pb_formulation} below.

% ===========================================================
%                 Problem formulation
% ===========================================================

\subsection{Goal-Conditioned Actor-Critic}
%\subsection{Problem formulation}
\label{subsec:pb_formulation}

We consider a discounted, infinite-horizon, goal-conditioned Markov decision process, with states $s \in \mathcal{S}$, goals $g \in \mathcal{G}$, actions $a \in \mathcal{A}$, reward function $r(s, a, g)$, dynamics $p(s' | s, a)$ and discount factor $\gamma$.
The objective of a goal-conditioned RL agent is to maximize the expected discounted return 
\begin{equation*}
    J(\pi) = \mathbb{E}_{g \sim \rho_g, \tau \sim d^\pi(.|g)} [\sum_{t} \gamma^t r(s_t, a_t, g)]
\end{equation*}
under the distribution 
\begin{equation*}
d^\pi(\tau | g) = \rho_0(s_0)\prod_t \pi(a_t|s_t, g)p(s_{t+1}|s_t, a_t) 
\end{equation*}
induced by the policy $\pi$ and the initial state and goal distribution.
The policy $\pi(.|s, g)$ in this work generates a distribution over continuous actions $a$ conditioned on state $s$ and goal $g$.
Many algorithms rely on the appropriate learning of the goal-conditioned action-value function $Q^\pi$ and the value function $V^\pi$
defined as
$Q^\pi(s, a, g) = \mathbb{E}_{s_0 = s, a_0=a, \tau \sim d^\pi(.|g)} [ \sum_t \gamma^t r(s_t, a_t, g)]$
and $V^\pi(s, g) = \mathbb{E}_{a \sim \pi(.|s,g)}  Q^\pi(s, a, g)$.

In this work we assume states and goals to co-exist in the same space, i.e. $\mathcal{S} = \mathcal{G}$, where each state can be considered as a potential goal.
Moreover, we set the reward $r$ to $-1$ for all actions until the policy reaches the goal.

We follow the standard off-policy actor-critic paradigm \cite{silver2014deterministic, heess2015learning, mnih2016asynchronous, fujimoto2018addressing,haarnoja2018soft}. Experience, consisting of single transition tuples $(s_t, a_t, s_{t+1}, g)$, is collected by the policy in a replay buffer $D$. 
Actor-critic algorithms maximize return by alternating between policy evaluation and policy improvement.
During the policy evaluation phase, a critic $Q^\pi(s, a, g)$ estimates the action-value function of the current policy $\pi$ by minimizing the Bellman error with respect to the Q-function parameters $\phi_k$:
\begin{equation} \label{eq:policy_evaluation}
    Q_{\phi_{k+1}} = \arg \min_\phi \frac{1}{2}
    \mathbb{E}_{
    (s_t, a_t, s_{t+1}, g) \sim D}[y_t -  Q_\phi(s_t, a_t, g)  ]^2
\end{equation}
with the target value
\begin{equation*}
    y_t = r(s_t, a_t, g) + \gamma \mathbb{E}_{a_{t+1} \sim \pi(.|s_t, g)} Q_{\phi_k} (s_{t+1}, a_{t+1}, g).
\end{equation*}
During the policy improvement phase, the actor $\pi$ is typically updated such that the expected value of the current Q-function $Q^\pi$, or alternatively the advantage $A^\pi(s, a, g) = Q^\pi(s, a, g) - V^\pi(s, g)$, under  $\pi$ is maximized:
\begin{equation} \label{eq:policy_improvement}
    \pi_{\theta_{k+1}} = \arg \max_\theta \mathbb{E}_{(s, g) \sim D, a \sim \pi_\theta(.|s,g) } [Q^\pi(s, a, g)].
\end{equation}

Reaching distant goals using delayed rewards may require expensive policy search. 
To accelerate the training, we propose to direct the policy search towards intermediate subgoals of a task. 
Our method learns to predict appropriate subgoals by the high-level policy $\pi^H$. 
The high-level policy is trained together with the policy $\pi$ as explained below.

% ===========================================================
%                 High-level policy training
% ===========================================================

\begin{figure}
\centering
\includegraphics[width=0.9\linewidth]{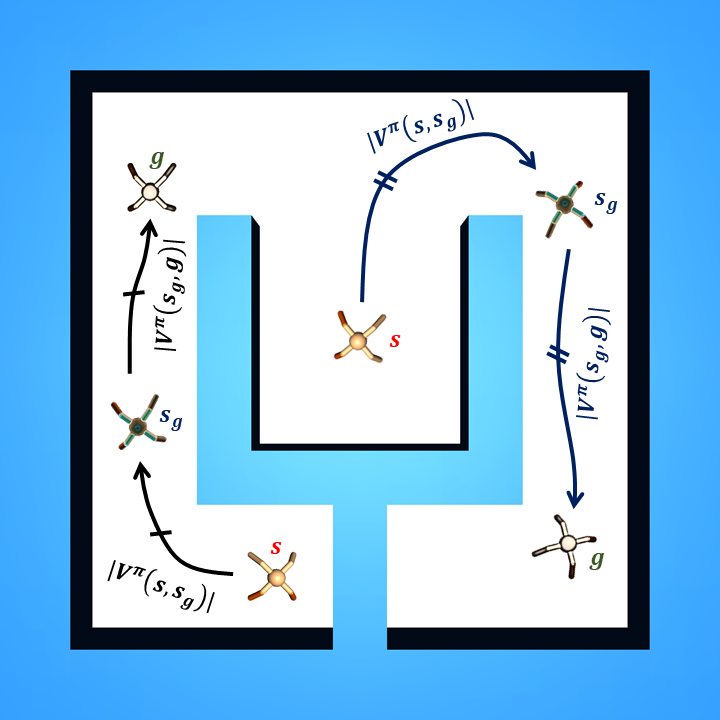}
\caption{
Given initial states $s$ and goal states $g$, our subgoals are states on the middle of the path from $s$ to $g$. We measure distances between states by the value function $|V^\pi(s_1, s_2)|$ corresponding to the current policy $\pi$. We obtain a distribution of subgoals from the high-level policy $s_g \sim \pi^H(.|s, g)$. We use subgoals {\em only} at the training to regularize and accelerate the policy search. 
}
\label{fig:ValueReachability}
\end{figure}

\subsection{High-Level Policy} \label{subsec:hp}
%\subsection{High-Level Policy \ec{Improvement}} \label{subsec:hp}

We would like to learn a high-level policy $\pi^H(.|s, g)$ 
%operating in the state space 
that predicts an appropriate distribution of imagined subgoals $s_g$ conditioned on valid states $s$ and goals $g$. Our high-level policy is defined in terms of the policy $\pi$ and relies on the goal-reaching capabilities of $\pi$ as described next.

\paragraph{Subgoal search with a value function.}

We note that our choice of the reward function $r=-1$ implies that the norm of the value function $V^\pi(s, g)$ corresponds to an estimate of the expected discounted number of steps required for the policy to reach the goal $g$ from the current state $s$.
We therefore propose to use $|V^\pi(s^i, s^j)|$ as a measure of the distance between any valid states $s^i$ and $s^j$. Note that this measure depends on the policy $\pi$ and evolves with the improvements of $\pi$ during training. 
%defined with respect to the current policy $\pi$.

Reaching imagined subgoals of a task should be easier than reaching the final goal. 
%Appropriate subgoals should be easier to reach, yet relevant for achieving the goal.
%As we later aim to incorporate subgoals into policy learning, 
To leverage this assumption for policy learning, we need to find appropriate subgoals. In this work we propose to define subgoals $s_g$ as midpoints on the path from the current state $s$ to the goal $g$, see Figure~\ref{fig:ValueReachability}. More formally, we wish $s_g$ (i)~to have equal distance from $s$ and $g$, and (ii)~to minimize the length of the paths from $s$ to $s_g$ and from $s_g$ to $g$.
We can find subgoals that satisfy these constraints by using our distance measure $|V^\pi(s^i, s^j)|$ and minimizing the following cost $C_\pi(s_g|s, g)$
\begin{equation}
    C_\pi(s_g|s, g) = \max \left ( |V^\pi(s, s_g)|, |V^\pi(s_g, g)| \right ).
    \label{eq:cost}
\end{equation}

However, naively minimizing the cost $C_\pi$ under the high-level policy distribution, i.e.
\begin{equation}
    \pi^H_{k+1} = \arg \min_{\pi^H} \mathbb{E}_{ (s, g) \sim D, s_g \sim \pi^H(.|s, g)} [C_\pi(s_g|s, g)],
    \label{eq:unregularized}
\end{equation}
may lead to undesired solutions where the high-level policy samples subgoals outside the valid state distribution $p_s(.)$. 
Such predictions may, for example, correspond to unfeasible robot poses, unrealistic images, or other adversarial states which have low distance from $s$ and $g$ according to $|V^\pi|$ but are unreachable in practice.
%that are misleadingly considered easily reachable by the RL agent.

\paragraph{Predicting valid subgoals.}
To avoid non-valid subgoals, we can additionally encourage the high-level policy to stay close to the valid state distribution $p_s(.)$ using the following KL-regularized objective:
\begin{equation} 
\label{eq:h_pi_equation}
\begin{split}
    \pi^H_{k+1} 
    &= \arg \max_{\pi^H} \mathbb{E}_{ (s, g) \sim D, s_g \sim \pi^H(.|s, g)} \left [ A^{\pi^H_k} (s_g|s, g) \right ] \\
    &\text{ s.t. } \DKL{\pi^H(.|s, g) }{p_s(.)} \leq \epsilon,
\end{split}
\end{equation}
where the advantage 
\begin{equation*}
A^{\pi^H_k}(s_g| s, g) = \mathbb{E}_{\hat{s_g} \sim \pi^H_k(.|s, g)} \left [C_\pi(\hat{s_g} |s, g) \right ]  - C_\pi(s_g |s, g)
\end{equation*}
 quantifies the quality of a subgoal $s_g$ against the high-level policy distribution.
The Kullback-Leibler divergence term in (\ref{eq:h_pi_equation}) requires an esimate of the density $p_s(.)$. While estimating the unknown $p_s(.)$ might be challenging, we can obtain samples from this distribution, for example, by randomly sampling states from the replay buffer $s_g \sim D$.

We propose instead to \textit{implicitly} enforce the KL constraint (\ref{eq:h_pi_equation}). 
We first note that the analytic solution to~(\ref{eq:h_pi_equation}) can be obtained by enforcing the Karush–Kuhn–Tucker conditions, for which the Lagrangian is:
\begin{equation*}
\begin{split}
    \mathcal{L}(\pi^H, \lambda) 
    &= \mathbb{E}_{s_g \sim \pi(.|s, g)} \left [A^{\pi^H_k}(s_g| s, g) \right ] \\
    &+ \lambda \left (\epsilon - \DKL{\pi(.|s, g)}{p_s(.)} \right ).
\end{split}
\end{equation*}
The closed form solution to this problem is then given by:
\begin{equation}
    \pi_\star^H(s_g |s, g) = \frac{1}{Z(s, g)} p_s(s_g) \exp \left (\frac{1}{\lambda} A^{\pi^H_k}(s_g| s, g) \right )
\end{equation}
with the normalizing partition function $Z(s, g) = \int p_s(s_g) \exp \left (\frac{1}{\lambda} A^{\pi^H_k}(s_g| s, g) \right )ds_g$ \cite{peters2010relative, rawlik2012stochastic, abdolmaleki2018maximum, abdolmaleki2018relative,  nair2020accelerating}.
We project this solution into the policy space by minimizing the forward KL divergence between our parametric high-level policy $\pi^H_\psi$ and the optimal non parametric solution $\pi_\star^H$:
\begin{equation}
\label{eq:h_pi}
\begin{split}
    \pi^H_{\psi_{k+1}} 
    &= \arg \min_\psi \mathbb{E}_{(s, g) \sim D} \DKL{\pi_\star^H(.|s, g)}{\pi^H_\psi(.|s, g)} \\
    = \arg &\max_\psi \mathbb{E}_{(s, g) \sim D, s_g \sim D}  \\
    & \left[\log \pi^H_\psi(s_g |s, g) \frac{1}{Z(s, g)} \exp \left (\frac{1}{\lambda} A^{\pi^H_k}(s_g| s, g) \right ) \right ],
\end{split}
\end{equation}
where $\lambda$ is a hyperparameter.
Conveniently, this policy improvement step corresponds to a weighted maximum likelihood with subgoal candidates obtained from the replay buffer randomly sampled among states visited by the agent in previous episodes. The samples are re-weighted by their corresponding advantage, implicitly constraining the high-level policy to stay close to the valid distribution of states. 
A more detailed derivation is given in Appendix \ref{ap:hpi}.

% ===========================================================
%        Policy regularization with imagined subgoals
% ===========================================================

\subsection{Policy Improvement with Imagined Subgoals}
\label{subsec:pi}

Our method builds on the following key insight.
%(Fig. \ref{fig:opener})
If we assume $s_g$ to be an intermediate subgoal on the optimal path from $s$ to $g$, then the optimal action for reaching $g$ from $s$ should be similar to the optimal action for reaching $s_g$ from $s$ (see Figure~\ref{fig:opener}).
We can formalize this using a KL constraint on the policy distribution, conditioned on goals $g$ and $s_g$:
\begin{equation*}
  \DKL{\pi(.|s, g)}{\pi(.|s, s_g)} \leq \epsilon.
\end{equation*}

We introduce the non-parametric prior policy $\pi^{prior}(.|s, g)$ as the distribution over actions that would be chosen by the policy $\pi$ for reaching subgoals $s_g \sim \pi^H(.|s, g)$ provided by the high-level policy.
Given a state $s$ and a goal $g$, we bootstrap the behavior of the policy at subgoals $s_g \sim \pi^H(.|s, g)$ (see Figure~\ref{fig:PriorPolicy}):
\begin{equation}
\label{eq:prior}
    \pi^{prior}_k(a |s, g) := \mathbb{E}_{s_g \sim \pi^H(.|s, g)} [\pi_{\theta_k'}(a|s, s_g)].
\end{equation}
As we assume the subgoals to be easier to reach compared to reaching the final goals, this prior policy provides a good initial guess to constrain the policy search to the most promising actions.

We then propose to leverage a policy iteration scheme with additional KL constraint to shape the policy behavior accordingly.
During the policy improvement step, in addition to maximizing the Q-function as in (\ref{eq:policy_improvement}), we encourage the policy to stay close to the prior policy through the KL-regularization:
\begin{equation}
\label{eq:pi}
\begin{split}
    \pi_{\theta_{k+1}} &= \arg \max_\theta
    \mathbb{E}_{(s, g) \sim D}  
    \mathbb{E}_{a \sim \pi_\theta(.|s, g)} \\
    &\left [ Q^\pi(s, a, g) - \alpha \DKL{\pi_\theta(.|s, g)}{\pi^{prior}_k(.|s, g)}
    \right ],
\end{split}
\end{equation}
where $\alpha$ is a hyperparameter.

In practice, we found that using an exponential moving average of the online policy parameters to construct the prior policy is necessary to ensure convergence:
\begin{equation}
\label{eq:movingaverage}
    \theta_{k+1}' = \tau \theta_{k} + (1-\tau) \theta_k',\text{ } \tau \in ]0, 1[.
\end{equation}
This ensures that the prior policy produces a more stable target for regularizing the online policy.

% ===========================================================
%        Algorithm Summary
% ===========================================================

\subsection{Algorithm Summary}
\label{subsec:algo}

\begin{algorithm}
    \caption{RL with imagined subgoals}
   \label{algo:algo}
\begin{algorithmic}
  \STATE Initialize replay buffer $D$
  \STATE Initialize $Q_\phi$, $\pi_\theta$, $\pi^H_\psi$
  \FOR {k = 1, 2, ...}
  \STATE Collect experience in $D$ using $\pi_\theta$ in the environment
  \STATE Sample batch $(s_t, a_t, r_t, s_{t+1}, g) \sim D$ with HER
  \STATE Sample batch of subgoal candidates $s_g \sim D$
  \STATE Update $Q_\phi$ using Eq. \ref{eq:policy_evaluation} \textit{\small (Policy Evaluation)}
  \STATE Update $\pi^H_\psi$ Using Eq. \ref{eq:h_pi} \textit{\small (High-Level Policy Improvement)}
  \STATE Update $\pi_\theta$ using Eq. \ref{eq:pi} \textit{\small (Policy Improvement with Imagined Subgoals)}
  \ENDFOR
\end{algorithmic}
\end{algorithm}

We approximate the policy $\pi_\theta$, the Q-function $Q_\phi$ and the high-level policy $\pi_\psi^H$
with neural networks parametrized by $\theta$, $\phi$ and $\psi$ respectively, and train them jointly using stochastic gradient descent.
The Q-function is trained to minimize the Bellman error (\ref{eq:policy_evaluation}), where we use an exponential moving average of the online Q-function parameters to compute the target value.
The high-level policy is a probabilistic neural network whose output parametrizes a Laplace distribution with diagonal variance $\pi^H_\psi(.|s, g) = \text{Laplace} (\mu^H_\psi(s, g), \Sigma^H_\psi(s, g))$ trained to minimize (\ref{eq:h_pi}).
The policy is also a probabilistic network parametrizing a squashed Gaussian distribution with diagonal variance $\pi^H_\psi(.|s, g) = \tanh \mathcal{N}(\mu_\theta(s, g), \Sigma_\theta(s, g))$ trained to minimize (\ref{eq:pi}).
Finally we can approximate $\pi_{prior}$ using  a Monte-Carlo estimate of (\ref{eq:prior}). 

The policy $\pi$ and the high-level policy $\pi^H$ are trained jointly.
As the policy learns to reach more and more distant goals, its value function becomes a better estimate for the distance between states.
This allows the high-level policy to propose more appropriate subgoals for a larger set of goals.
In turn, as the high-level policy improves, imagined subgoals offer a more relevant supervision to shape the behavior of the policy.
This virtuous cycle progressively extends the policy horizon further and further away, allowing more complex tasks to be solved by a single policy.

The full algorithm is summarized in Algorithm \ref{algo:algo}.
We use hindsight experience replay \cite{andrychowicz2017hindsight} (HER) to improve learning from sparse rewards.
Additional implementation details are given in Appendix \ref{ap:implem_details}.

%% file: Sections/Experiments.tex
\begin{figure}[t!]
\begin{subfigure}{0.5\columnwidth}
\centering
\includegraphics[width=1.0\linewidth]{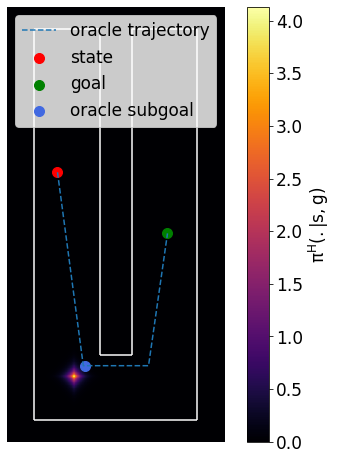}
\end{subfigure}%
\begin{subfigure}{0.5\columnwidth}
\centering
\includegraphics[width=1.0\linewidth]{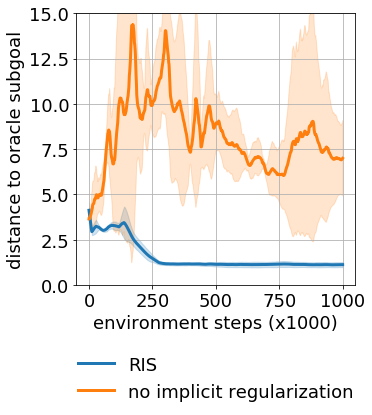}
\end{subfigure}%
\caption{
(Left) Heatmap of the subgoal distribution obtained with our high-level policy and the oracle subgoal for a given state and goal for the ant U-maze environment.
(Right) Distance between oracle subgoals and subgoals predicted by the high-level policy for RIS and RIS without implicit regularization. 
The dimensions of the space are $7.5 \times 18$ units and the ant has a radius of roughly 0.75 units.
}
\vspace{-0.3cm}
\label{fig:oracle_subgoal}
\end{figure}

\begin{figure}[t!]
\begin{subfigure}{0.5\columnwidth}
\centering
\includegraphics[width=1.0\linewidth]{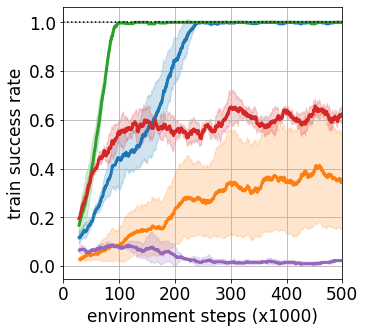}
\end{subfigure}%
\begin{subfigure}{0.5\columnwidth}
\centering
\includegraphics[width=1.0\linewidth]{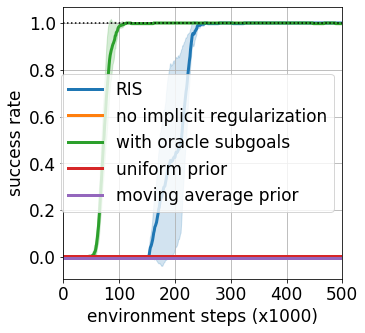}
\end{subfigure}%
\caption{
%Comparison to a number of variants on the Ant U-Maze environment:
Ablation of our method on the Ant U-Maze environment:
simple priors that do not incorporate subgoals (\textit{uniform prior, moving average prior});
ignoring the effect of out-of-distribution subgoal predictions (\textit{no implicit regularization}); and using oracle subgoals (\textit{with oracle subgoals}). Left: success rate on all configurations throughout training. Right: success rate on the test configurations.
}
\vspace{-0.3cm}
\label{fig:ablation}
\end{figure}

\section{Experiments} \label{section:experiments}

In this section we first introduce our experimental setup in Section \ref{sec:exp}.
Next, we ablate various design choices of our approach in Section \ref{subsec:ablation}. We, then, compare RIS to prior work in goal-conditioned reinforcement learning in Section \ref{subsec:comparison}.

\subsection{Experimental Setup}
\label{sec:exp}

\begin{figure*}[ht]
\centering
\vspace{.2cm}
\begin{subfigure}{0.25\textwidth}
\includegraphics[width=1.0\linewidth]{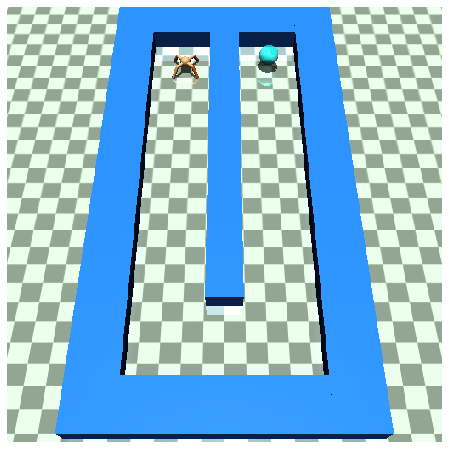}
\vspace{-.25cm}\\
\includegraphics[width=1.0\linewidth]{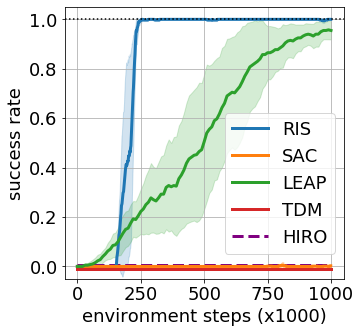}
\caption{U-shaped maze}
\end{subfigure}%
\begin{subfigure}{0.25\textwidth}
\includegraphics[width=1.0\linewidth]{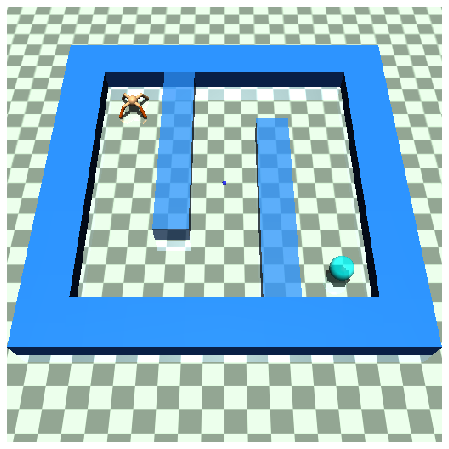}
\vspace{-.25cm}\\
\includegraphics[width=1.0\linewidth]{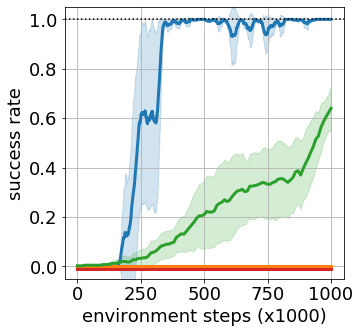}
\caption{S-shaped maze}
\end{subfigure}%
\begin{subfigure}{0.25\textwidth}
\includegraphics[width=1.0\linewidth]{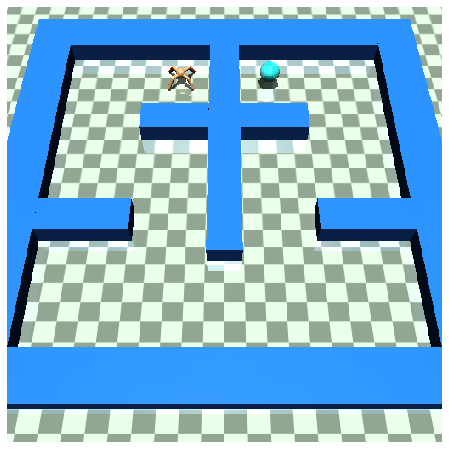}
\vspace{-.25cm}\\
\includegraphics[width=1.0\linewidth]{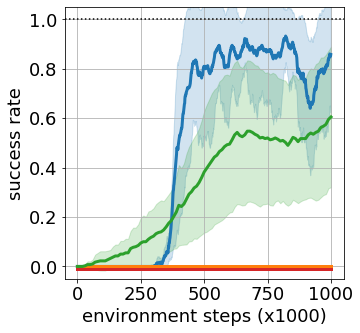}
\caption{$\Pi$-shaped maze}
\end{subfigure}%
\begin{subfigure}{0.25\textwidth}
\includegraphics[width=1.0\linewidth]{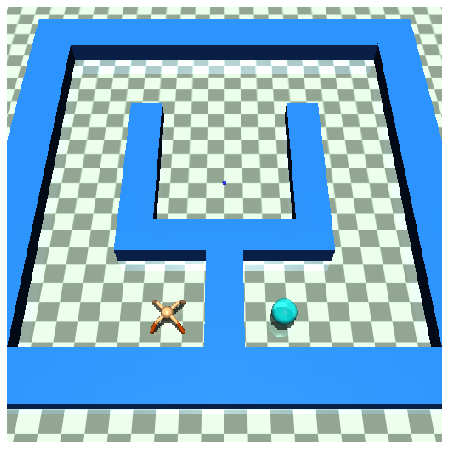}
\vspace{-.25cm}\\
\includegraphics[width=1.0\linewidth]{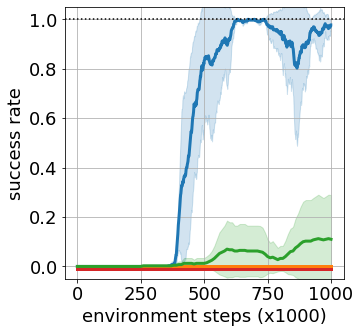}
\caption{$\omega$-shaped maze}
\end{subfigure}
\caption{
Comparison of RIS to several state-of-the-art methods (bottom row) on 4 different ant navigation tasks.
We evaluate the success rate of the agent on the challenging configurations illustrates in the top row, where the ant is located at the initial state and the desired goal location is represented by a cyan sphere.
}
%\vspace{-.2cm}
\label{fig:antmaze_navigation}
\end{figure*}

\paragraph{Ant navigation.}
We evaluate RIS on a set of ant navigation tasks of increasing difficulty, each of which requires temporally extended reasoning.
In these environments, the agent observes the joint angles, joint velocity, and center of mass of a quadruped ant robot navigating in a maze.
We consider four different mazes: a U-shaped maze, a S-shaped maze, a $\Pi$-shaped maze and a $\omega$-shaped maze illustrated in Figure~\ref{fig:antmaze_navigation}.
The obstacles are unknown to the agent.

During training, initial states and goals are uniformly sampled and the agents are trained to reach any goal in the environment.
We evaluate agents in the most extended temporal settings representing the most difficult configurations offered by the environment (see Figure~\ref{fig:antmaze_navigation}).
We assess the success rate achieved by these agents, where we define success as the ant being sufficiently close to the goal position measured by x-y Euclidean distance.

\paragraph{Vision-based robotic manipulation.}
We follow the experimental setup in~\cite{nasiriany2019planning} and
also consider a vision-based robotic manipulation task where an agent controls a 2 DoF robotic arm from image input and must manipulate a puck positioned on the table (Figure \ref{fig:PushAndReach:Illustration}).
We define success as the arm and the puck being sufficiently close to their respective desired positions.
During training, the initial arm and puck positions and their respective desired positions are uniformly sampled whereas, at test time, we evaluate agents on temporally extended configurations.

These tasks are challenging because they require temporally extended reasoning on top of complex motor control.
Indeed, a greedy path towards the goal cannot solve these tasks. 
We train the agents for 1 million environment steps and average the results over 4 random seeds.
Additional details about the environments are given in Appendix \ref{ap:environments}.

%\subsubsection{Alternative methods} \label{subsubsec:baselines}
\paragraph{Alternative methods.}
We compare our approach, RIS, to off-policy reinforcement learning methods for goal-reaching tasks.
We consider Soft Actor-Critic (SAC) \cite{haarnoja2018soft} with HER, which trains a policy from scratch by maximizing the entropy regularized objective using the same sparse reward as ours.
We also compare to Temporal Difference Models (TDM) \cite{pong2018temporal} which trains horizon-aware policies operating under dense rewards in the form of distance to the goal.
We chose to evaluate TDMs with a long policy horizon of 600 steps due to the complexity of considered tasks.
Furthermore, we compare to Latent Embeddings for Abstracted Planning
(LEAP) \cite{nasiriany2019planning}, a competitive approach for these environments,
which uses a sequence of subgoals that a TDM policy must reach one after the other during inference. 

We re-implemented SAC, TDM and LEAP and validated our implementations on the U-shaped ant maze and vision-based robotic manipulation environments.

On the U-shaped Ant maze environment, we additionally report results of HIRO \cite{nachum2018data}, a hierarchical reinforcement learning method with off-policy correction, after 1 million environment steps. The results were copied from Figure 12 in  \cite{nasiriany2019planning}.

We provide additional details about the hyperparameters used in our experiments in Appendix \ref{ap:hyperparameters}.

% ===============================================================================
%                    Ablative Analysis
% ===============================================================================

\subsection{Ablative Analysis} \label{subsec:ablation}

We use the Ant U-maze navigation task to conduct ablation studies to validate the role of our design choices.

\begin{figure*}[ht]
\vspace{3mm}
\centering
\begin{subfigure}{1.05\columnwidth}
\centering
\begin{subfigure}{0.5\columnwidth}
\includegraphics[width=1.1\linewidth]{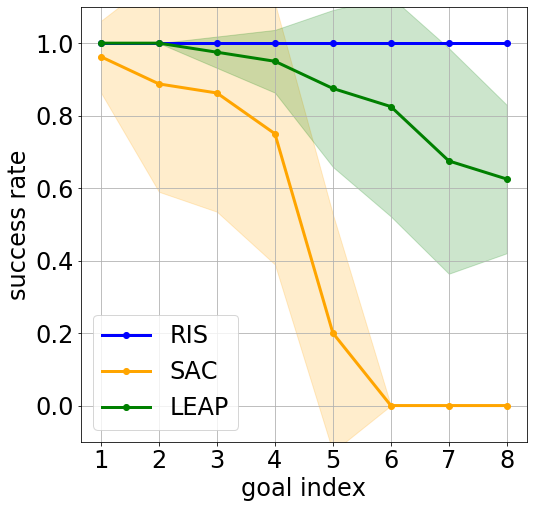}
\end{subfigure}%
\begin{subfigure}{0.5\columnwidth}
\centering
\includegraphics[width=0.8\linewidth]{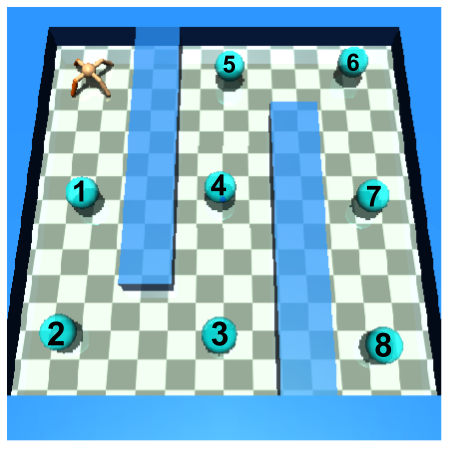}
\end{subfigure}%
\caption{S-shaped Maze}
\end{subfigure}%
\begin{subfigure}{1.05\columnwidth}
\centering
\begin{subfigure}{0.5\columnwidth}
\includegraphics[width=1.1\linewidth]{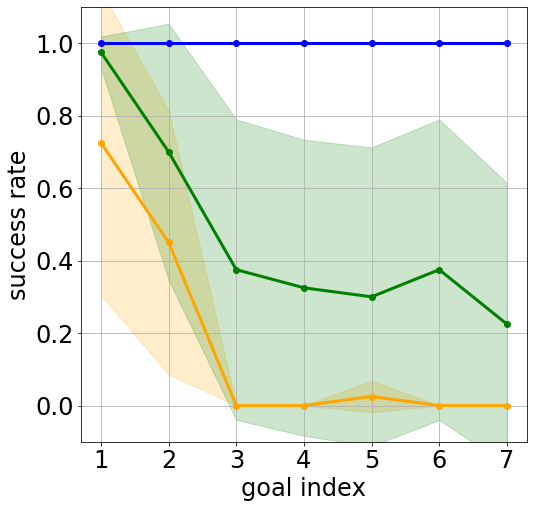}
\end{subfigure}%
\begin{subfigure}{0.5\columnwidth}
\centering
\includegraphics[width=0.8\linewidth]{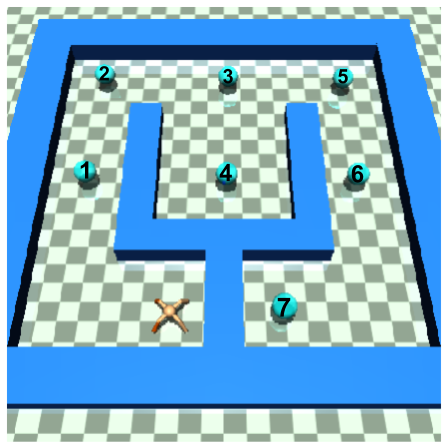}
\end{subfigure}%
\caption{$\omega$-shaped Maze}
\end{subfigure}%
\caption{Comparison of different methods for increasingly more difficult tasks on the S-shaped and $\omega$-shaped ant maze. The goals with increasing complexity are numbered starting with 1, where 1 is the closest goal. 
}
\label{fig:HorizonMazes}
\end{figure*}

\noindent{\bfseries Evaluating imagined subgoals.} 
%\paragraph{Evaluating imagined subgoals.} 
To evaluate the quality of the subgoals predicted by our high-level policy, we introduce an oracle subgoal sampling procedure.
We plan an oracle trajectory, which corresponds to a point-mass agent navigating in this maze and does not necessarily correspond to the optimal trajectory of an ant.
Oracle subgoals correspond to the midpoint of this trajectory between state and goal in x-y location (Figure \ref{fig:oracle_subgoal} left).
 
Figure \ref{fig:oracle_subgoal} (left) shows the subgoal distribution predicted by a fully-trained high-level policy for a state and goal pair located at opposite sides of the U-shaped maze. We can observe that its  probability mass is close to the oracle subgoal.

To quantitatively evaluate the quality of imagined subgoals, we measure the x-y Euclidean distance between oracle subgoals and subgoals sampled from the high-level policy throughout training for a set of fixed state and goal tuples randomly sampled in the environment.
Figure \ref{fig:oracle_subgoal} (right) shows that RIS successfully learns to find subgoals that are coherent with the oracle trajectory during training, despite not having prior knowledge about the environment.

We also assess the importance of using the implicit regularization scheme presented in Section \ref{subsec:hp} which discourages high-level policy predictions to lie outside of the distribution of valid states.
We compare our approach against naively optimizing subgoals without regularization (i.e.\,directly optimize (\ref{eq:unregularized})).
Figure \ref{fig:oracle_subgoal} (right) shows that without implicit regularization, the predicted subgoals significantly diverge  from oracle subgoals in x-y location during training.
As result, imagined subgoals are not able to properly guide policy learning to solve the task, see Figure~\ref{fig:ablation}.

\noindent{\bfseries Prior policy with imagined subgoals.}
%\paragraph{Prior policy with imagined subgoals.}
We evaluate the importance of incorporating imagined subgoals into the prior policy by  comparing to a number of variants.
To disentangle the effects of imagined subgoals from the actor-critic architecture used by RIS, we first replace our prior policy with simpler choices of prior distributions that do not incorporate any subgoals:
(i) a uniform prior over the actions $\pi^{prior}_k = \mathcal{U} (\mathcal{A})$, which is equivalent to SAC without entropy regularization during policy evaluation, and 
(ii) a parametric prior policy that uses an exponential moving average of the online policy weights  $\pi^{prior}_k = \pi_{\theta'_k}$.
Figure~\ref{fig:ablation} shows that, while they can learn goal-reaching behaviors on many configurations encountered during training (left), neither of these variants are able to solve the Ant U-maze environment in its most difficult setting (right). 
We also observe that the agent with a moving average action prior fails to learn (left).
This highlights the benefits of incorporating subgoals into policy learning.

Finally, we propose to incorporate the oracle subgoals into our prior policy.
We replace the subgoal distributions predicted by our high-level policy with Laplace distributions centered around oracle subgoals.
Results in Figure \ref{fig:ablation} show that RIS with oracle subgoals learns to solve the U-shaped ant maze environment faster than using a high-level policy simultaneously trained by the agent.
This experiment highlights the efficiency of our approach to guide policy learning with appropriate subgoals:
if we have access to proper subgoals right from the beginning of the training, our approach could leverage them to learn even faster.
However, such subgoals are generally not readily available without prior knowledge about the environment. 
Thus, we introduce a high-level policy training procedure which determines appropriate subgoals without any supervision.

% ===============================================================================
%                    Comparison to prior works
% ===============================================================================

\subsection{Comparison to the State of the Art} \label{subsec:comparison}

\noindent{\bfseries Ant navigation.}
Figure \ref{fig:antmaze_navigation} compares RIS to the  alternative methods introduced in section \ref{sec:exp} for the four ant navigation environments. 
For all considered mazes RIS significantly outperforms prior methods in terms of sample efficiency,
often requiring less than 500 thousand environment interactions to solve the mazes in their most challenging initial state and goal configurations.

LEAP makes progress on these navigation tasks, but requires significantly more environment interactions.
This comparison shows the effectiveness of our approach, which uses subgoals to guide the policy rather than reaching them sequentially as done by LEAP. 
While SAC manages to learn goal-reaching behaviors, as we will see later in Figure \ref{fig:HorizonMazes}, it fails to solve the environments in their most challenging configurations.
The comparison to SAC highlights the benefits of using our informed prior policy compared to methods assuming a uniform action prior.
On the U-shaped maze environment, HIRO similarly fails to solve the task within one million environment interactions. 
Furthermore, we observe that TDMs fails to learn due to the sparsity of the reward.

Figure \ref{fig:HorizonMazes} evaluates how SAC, LEAP and RIS perform for varying task horizons in the S-shaped and $\omega$-shaped mazes.
Starting from an initial state located at the edge of the mazes, 
we sample goals at locations which require an increasing number of environment steps to be reached.
Figure \ref{fig:HorizonMazes} reports results for RIS, LEAP and SAC after having been trained for 1 million steps.
While the performances of LEAP and SAC degrades as the planning horizon increases, RIS consistently solves configurations of increasing complexity. 

These results demonstrate that RIS manages to solve complex navigation tasks despite relying on a flat policy at inference. 
In contrast, LEAP performs less well, despite  a significantly more expensive planning of subgoals during inference. 

\begin{figure}[ht]
\vspace{4mm}
\centering
\begin{subfigure}[t]{0.45\columnwidth}
\includegraphics[width=1.0\linewidth]{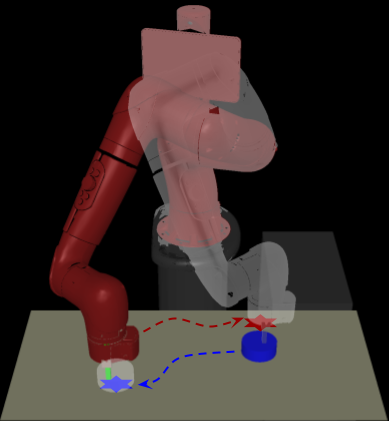}
\caption{Illustration of the robotic manipulation task}
\label{fig:PushAndReach:Illustration}
\end{subfigure}%
\begin{subfigure}[t]{0.55\columnwidth}
\includegraphics[width=1.0\linewidth]{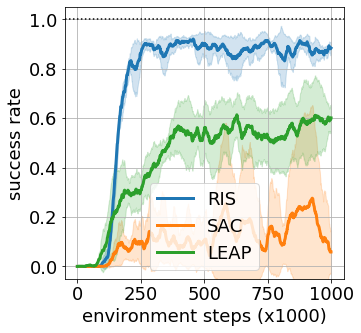}
\caption{Comparison to prior works}
\label{fig:PushAndReach:Curves}
\end{subfigure}%
\caption{
Robotic manipulation environment: (a) illustration of the task; (b) results of our method compared to LEAP and SAC.
}
\label{fig:PushAndReach}
\end{figure}

\noindent{\bfseries Vision-based robotic manipulation.}
While the ant navigation experiments demonstrate the performances of RIS on environments with low-dimension state spaces, we also show how our method can be applied to vision-based robotic manipulation tasks.
Our approach takes images of the current and desired configurations as input. Input images are passed through an image encoder, a convolutional neural network shared between the policy, high-level policy and Q-function.
The encoder is only updated when training the Q-function during policy evaluation and is fixed otherwise during policy improvement and high-level policy improvement.
Instead of generating subgoals in the high-dimensional image space, the high-level policy therefore operates in the learned compact image representation of the encoder.
Following recent works on reinforcement learning from images, we augment image observations with random translations~\cite{kostrikov2020image,
laskin2020reinforcement}.
We found that using such data augmentation was important for training image-based RIS and SAC policies.
Moreover, we found that using a lower learning rate for the policy was necessary to stabilize training.
Additional implementation details on the image encoding are given in Appendix~\ref{ap:implem_details}.

We compare our approach against LEAP and SAC in Figure~\ref{fig:PushAndReach:Curves}.
RIS achieves a higher success rate than LEAP whereas SAC fails most of the time to solve the manipulation task consistently enough in the temporally extended configuration used for evaluation on the vision-based robotic manipulation task.
Moreover, RIS and SAC only requires a single forward pass through their image encoder and actor network at each time step when interacting with the environment, whereas LEAP depends in addition upon an expensive planning of image subgoals. 

\begin{figure}[t]
\vspace{4mm}
\centering
\includegraphics[width=1.0\linewidth]{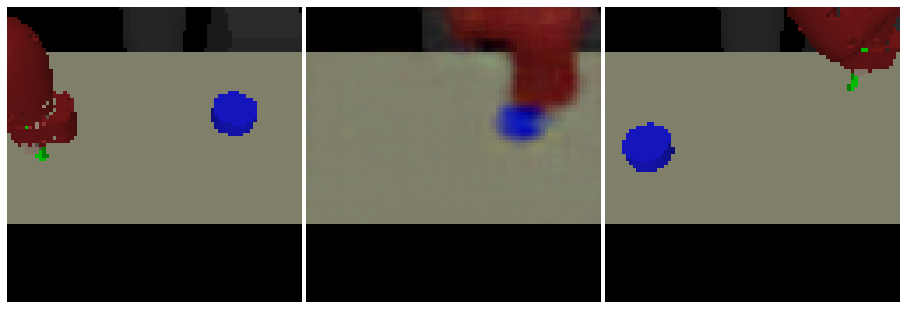}

\includegraphics[width=1.0\linewidth]{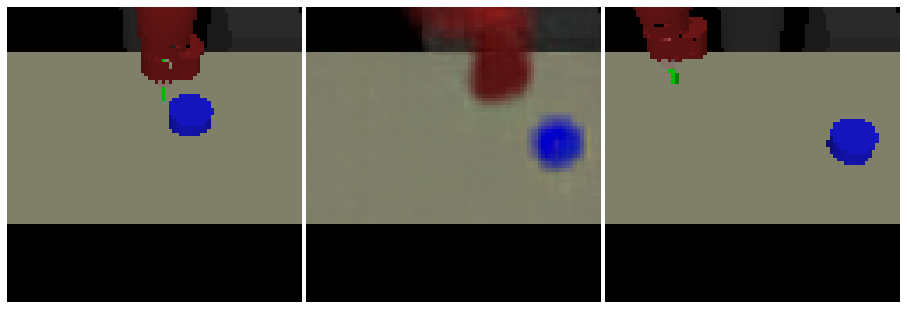}
\caption{Image reconstruction of an imagined subgoal (middle) given the current state (left) and the desired goal (right) on a temporally extended configuration used for evaluation (top) and a random configuration (bottom).}
%\vspace{-.3cm}
\label{fig:PushAndReach:ImaginedSubgoal}
\end{figure}

Figure~\ref{fig:PushAndReach:ImaginedSubgoal}
visualizes the imagined subgoals of the high-level policy.
Once the RIS agent is fully trained, we separately train a decoder to reconstruct image observations from their learned representations.
Given observations of the current state and the desired goal, we then  predict the representation of an imagined subgoal with the high-level policy and generate the corresponding image using the decoder.
Figure~\ref{fig:PushAndReach:ImaginedSubgoal} shows that subgoals predicted by the high-level policy are natural intermediate states halfway to the desired goal on this manipulation task.
For example, for the test configuration (Figure~\ref{fig:PushAndReach:ImaginedSubgoal} top), the high-level policy prediction  corresponds to a configuration where the arm has reached the right side of the puck and is pushing it towards its desired position.

Additional reconstructions of imagined subgoals for different initial state and goal configurations in the robotic manipulation environment are given in Appendix~\ref{ap:additionalresults}.

%% file: Sections/Conclusion.tex
\section{Conclusion}

We introduced RIS, a goal-conditioned reinforcement learning method that imagines possible subgoals in a self-supervised fashion and uses them to facilitate training.
We propose to use the value function of the goal-reaching policy to train a high-level policy operating in the state space.
We then use imagined subgoals to define a prior policy and incorporate this prior into policy learning.
Experimental results on challenging simulated navigation and vision-based manipulation environments show that
our proposed method greatly accelerates learning of temporally extended tasks and outperforms competing approaches. 

While our approach makes use of subgoals to facilitate policy search, future work could explore how to use them to obtain better Q-value estimates.
Future work could also improve exploration by using  imagined subgoals to encourage the policy to visit all potential states. 

\section{Acknowledgements}

This work was funded in part by the French government under management of Agence Nationale de la Recherche as part of the ”Investissements d’avenir” program, reference ANR19-P3IA-0001 (PRAIRIE 3IA Institute) and by Louis Vuitton ENS Chair on Artificial Intelligence.

%% file: Appendix/Derivation.tex
\section{High-level Policy Derivation}\label{ap:hpi}

The following derivation is similar to \cite{abdolmaleki2018maximum} and \cite{nair2020accelerating}.
We consider the subgoal advantage function $A^{\pi^H}$, some distribution over valid states $p_s(s)$ and the distribution over states and goals $\mu(s, g)$ given by previously collected experience and experience relabelling.
During high-level policy improvement, we maximize the following objective function, where expectations have been replaced by integrals:
\begin{equation}
\label{eq:objfun}
\begin{split}
    \pi^H_{k+1} &= \arg \max_{\pi} \iint \mu(s, g)\int \pi^H (s_g |s, g) A^{\pi^H}(s_g |s, g)ds_gdsdg \\
     \text{s.t. } \iint \mu(s, g) & \DKL{\pi^H(.|s, g)}{p_s(.)} dsdg \leq
     \epsilon,
     \iint  \mu(s, g) \int \pi^H(s_g|s, g)ds_gdsdg = 1
\end{split}
\end{equation}

The Lagrangian is:
\begin{equation}
\begin{split}
    \mathcal{L}(\pi^H, \lambda, \eta) &=
    \iint \mu(s, g) \int \pi^H(s_g|s, g) A^{\pi^H}(s_g |s, g)ds_gdsdg \\
    &+ \lambda \left(\epsilon - \iint \mu(s, g) \int \pi^H(s_g|s, g) \log \frac{\pi^H(s_g|s, g)}{p_s(s_g)}ds_gdsdg \right) \\
    &+ \eta\left(1 - \iint  \mu(s, g) \int \pi^H(s_g|s, g)ds_gdsdg\right).
\end{split}
\end{equation}
Differentiating with respect to $\pi^H$ yields
\begin{equation}
    \frac{\partial \mathcal{L}}{\partial\pi^H} =  A^{\pi^H}(s_g |s, g) - \lambda \log \pi^H(s_g |s, g) + \lambda \log p_s(s_g)  + \eta - \lambda.
\end{equation}
Setting this expression to zero, we get
\begin{equation}
    \pi^H(s_g |s, g) = p_s(s_g) \exp \left ( \frac{A^{\pi^H}(s_g |s, g)}{\lambda} \right ) \exp \left ( - \frac{\lambda - \eta}{\lambda} \right ).
\end{equation}
As the second constraint in (\ref{eq:objfun}) must sum to 1, the last exponential is a normalizing factor.
This gives the closed form solution:
\begin{equation}
    \pi_\star^H(s_g |s, g) = \frac{1}{Z(s, g)}p_s(s_g) \exp \left ( \frac{A^{\pi^H}(s_g |s, g)}{\lambda} \right )
\end{equation}
with the normalizing partition function
\begin{equation}
    Z(s, g) = \int p_s(s_g) \exp \left ( \frac{1}{\lambda} A^{\pi^H}(s_g |s, g) \right )ds_g.
\end{equation}

We next project the closed form solution into the space of parametric policies by minimizing the reverse KL divergence between our parametric high-level policy $\pi^H_\psi$ and the optimal non-parametric solution $\pi^H_\star$:

\begin{equation}
\begin{split}
    \pi^H_{\psi_{k+1}} 
    &= \arg \min_{\psi} \mathbb{E}_{ (s, g) \sim \mu(.)} \left [ \DKL{\pi_\star^H ( . |s, g) }{ \pi_\psi^H ( . |s, g) } \right] \\
    &= \arg \max_{\psi} \mathbb{E}_{ (s, g) \sim \mu(.)} \mathbb{E}_{s_g \sim \pi_\star^H ( . |s, g)} \left [ \log \pi^H_\psi (s_g |s, g) \right ] \\
    &= \arg \max_{\psi} \mathbb{E}_{ (s, g) \sim \mu(.),  s_g \sim p_s(.)} \left [ \log \pi_\psi^H (s_g |s, g) \frac{1}{Z(s, g)} \exp \left( \frac{ 1 }{ \lambda } A^{\pi^H}(s_g | s, g ) \right ) \right]
\end{split}
\label{eq_ap:h_pi}
\end{equation}

We choose $p_s$ to be the distribution of states given by experience collected in the replay buffer $D$,
such that high-level policy improvement corresponds to a weighted maximum-likelihood where subgoal candidates $s_g$ are randomly sampled from $D$ among states visited by the agent in previous episodes.

%% file: Appendix/Implementation_details.tex
\section{Implementation Details}
\label{ap:implem_details}

\paragraph{Actor-Critic.}
Our implementation of the actor-critic algorithm is based on 
Soft Actor-Critic \cite{haarnoja2018sacapps}, where we remove the entropy term during policy evaluation and replace the entropy term by the KL divergence between policy and our prior policy during policy improvement.
The policy is a neural network that parametrizes the mean and diagonal covariance matrix of a squashed Gaussian distribution $\pi_\theta(.|s, g) = \tanh \mathcal{N}(\mu_\theta(s, g), \Sigma_\theta(s, g))$.
We train two seperate Q-networks with target networks and take the minimum over the two target values to compute the bootstrap value.
The target networks are updated using an exponential moving average of the online Q parameters:
$\phi_{k+1}' = \tau \phi_k + (1-\tau) \phi_k'$.

\paragraph{High-level policy training.}
The high-level policy is a neural network that outputs the mean and diagonal covariance matrix of a Laplace distribution $\pi^H_\psi(.|s, g) = \text{Laplace}(\mu_\psi(s, g), \Sigma_\psi(s, g))$.
Following \cite{nair2020accelerating}, instead of estimating the normalizing factor $Z(s, g)$ in (\ref{eq_ap:h_pi}),
we found that computing the weights as softmax of the advantages over the batch leads to good results in practice.
During the high-level policy improvement, we found that clipping the value function between $-100$ and $0$, which corresponds to the expected bounds given our choice of reward function and discount factor, stabilizes the training slightly.

\paragraph{KL divergence estimation.}
We use an exponentially moving average of the policy weights instead of the weights of the current policy to construct the prior policy $\pi_{prior}$: $\theta_{k+1}' = \tau \theta_k + (1 - \tau)\theta_{k}'$ 
with the same smoothing coefficient $\tau$ as the one used for the Q function.
We estimate the prior density using the following Monte-Carlo estimate:
\begin{equation}
\label{eq:prior_mc}
   \log \pi^{prior}(a |s, g) \approx \log \left [ \frac{1}{I} \sum_i \pi_{\theta'} (a |s, s_g^i ) + \epsilon \right ], (s_g^i) \sim \pi^H_\psi(.|s, g),
\end{equation}
where $\epsilon > 0$ is a small constant to avoid large negative values of the prior log-density.
We use $I=10$ samples to estimate $\pi^{prior}(a|s, g)$.
We also use a Monte-Carlo approximation to estimate the KL-divergence term in Equation (9) of the submission:
%(\ref{eq:pi}):
\begin{equation}
\begin{split}
    \DKL{\pi_\theta(.|s, g)}{\pi^{prior}(.|s, g)} 
    &= \mathbb{E}_{a \sim \pi(.|s, g)} [ \log \pi_\theta(a|s, g) - \log \pi^{prior}(a|s, g)] \\
    &\approx \frac{1}{N} \sum_n \left [ \log \pi_\theta(a_n|s, g) - \log \pi^{prior}(a_n|s, g) \right ] \\
    \text{with } (a_n)_{n=1,..,N} &\sim \pi_\theta(.|s, g).
\end{split}
\end{equation}
Following SAC \cite{haarnoja2018soft}, we use $N=1$, plug the estimate (\ref{eq:prior_mc}) and use the reparametrization trick to backpropagate the KL divergence term to the policy weights \cite{haarnoja2018soft}.

\paragraph{Experience relabelling.}
In all of our experiments we use Hindsight Experience Replay \cite{andrychowicz2017hindsight}.
We use the same relabelling strategy as \cite{nair2018visual} and \cite{nasiriany2019planning} and relabel the goals in our minibatches as follows:
\vspace{-.3cm}
\begin{itemize}
    \item $20\%$: original goals from collected trajectories,\vspace{-.2cm}
    \item $40\%$: randomly sampled states from the replay buffer, trajectories,\vspace{-.2cm}
    \item $40\%$: future states along the same collected trajectory.
\end{itemize}

\paragraph{Vision based environments}
On the vision-based robotic manipulation tasks, input images are passed through an image encoder shared between the policy, high-level policy and Q-function.
Both states, goals and subgoals are encoded using the same encoder network.
The encoder is updated during policy evaluation, where we only update the representation of the current state images whereas the representations of desired goal images, next state images and subgoal image candidates are kept fixed.
We augment the observations with random translations by translating the $84 \times 84$ image within a $100 \times 100$ empty frame \cite{laskin2020reinforcement, kostrikov2020image}.

%% file: Appendix/Environments.tex
\section{Environments}
\label{ap:environments}

We adopt Ant Navigation environments provided by the code repository of~\cite{nasiriany2019planning}.
In these environments, a 4-legged ant robot must learn to navigate in various mazes.
The state includes the position, orientation, the joint angles and the velocities of these components. 
The ant has a radius of roughly 0.75 units.
We consider that the goal is reached if the x-y position of the ant is within 0.5 units of its desired location in Euclidean distance.
The agent receives a reward $r(s, a, g) = -1$ for all actions until the goal is reached.
The dimensions of the space for the U-shaped maze are $7.5 \times 18$ units.
For the S-shaped maze, the dimensions are $12 \times 12$.
For the $\Pi$-shaped and $\omega$-shaped mazes, the dimensions are $16 \times 16$ units.
The walls are $1.5$ units thick.
During training, initial states and goals are uniformly sampled anywhere in the empty space of the environment.
At test time, we evaluate the agent on challenging configurations that require temporally extended reasoning, as illustrated in Figure~6 of the submission. 

For the robotic manipulation task, we use the same image-based environment as in \cite{nasiriany2019planning}. 
In this task, the agent operates an arm robot via 2D position control and must manipulate a puck.
The agent observes a  $84 \times 84$ RGB image showing a top-down view of the scene. 
The dimension of the workspace are  $40 \text{cm} \times 20 \text{cm}$ and the puck has a radius of $4 \text{cm}$. 
We consider that the goal is achieved if both the arm and the puck are within $5 \text{cm}$ of their respective target positions.
During training, the initial arm and puck positions and their respective desired positions are uniformly sampled in the workspace.
At test time, we evaluate the policy on a hard configuration which requires temporally extended reasoning:
the robot must reach across the table to a corner where the puck is located, move its arm around the puck, pull the puck to a different corner of the table, and reach again the opposite corner.

%% file: Appendix/Hyperparameters.tex
\section{Hyperparameters}
\label{ap:hyperparameters}

Table \ref{table:RISSAChyperparameters} lists the hyperparameters used for the RIS.
We use Adam optimizer and report results after one million interactions with the environment.
For SAC, following \cite{haarnoja2018sacapps}, we automatically tune the entropy of the policy to match the target entropy of $-\text{ dim} (\mathcal{A})$.

\begin{table}[h]
\mbox{}\vspace{-.7cm}\\
\caption{Hyper-parameters for RIS and SAC.}
\label{table:RISSAChyperparameters}
\vskip 0.15in
\centering
\begin{small}
\begin{tabular}{ l | l | l }
\toprule
Hyper-parameter & Ant Navigation & Robotic Manipulation \\ 
\midrule
Q hidden sizes & $[256, 256]$ & $[256, 256]$ \\ 
Policy hidden sizes & $[256, 256]$ & $[256, 256]$  \\ 
High-level policy hidden sizes & $[256, 256]$ & $[256, 256]$\\ 
Hidden activation functions  &  ReLU &  ReLU\\ 
Batch size & $2048$ & $256$ \\
Training batches per environment step & $1$ & $1$\\
Replay buffer size & $1 \times 10^6$ & $1 \times 10^5$ \\
Discount factor $\gamma$ & $0.99$ & $0.99$ \\
polyak for target networks $\tau$ & $5\times 10^{-3}$ & $5\times 10^{-3}$ \\
$\epsilon$ & $1 \times 10^{-16}$ & $1 \times 10^{-4}$ \\
Critic learning rate & $1 \times 10 ^{-3}$ & $1 \times 10 ^{-3}$ \\
Policy learning rates & $1 \times 10 ^{-3}$ & $1 \times 10 ^{-4}$ \\
High-level policy learning rate & $1 \times 10^{-4}$ & $1 \times 10^{-4}$ \\
$\alpha$ & $0.1$ & $0.1$ \\
$\lambda$ & $0.1$ & $0.1$ \\
\bottomrule
\end{tabular}
\end{small}
\end{table}

In the vision based environment, our image encoder is a serie of convolutional layers with kernel sizes $[3, 3, 3, 3]$, strides $[2, 2, 2, 1]$, channel sizes $[32, 32, 32, 32]$ and $ReLU$ activation functions followed by a fully-connected layer with output dimension $16$.

\begin{table}[h]
\mbox{}\vspace{-.7cm}\\
\caption{Environment specific hyper-parameters for LEAP}
\label{table:LEAPhyperparameters}
\vskip 0.15in
\centering
\begin{small}
\begin{tabular}{ l | l |l | l | l | l }
\toprule
Hyperparameter & U-shaped maze & S-shaped maze & $\Pi$-shaped maze & $\omega$-shaped Maze & Robotic Manipulation  \\ 
\midrule
TDM policy horizon & 50 & 50 & 75 & 100 & 25  \\ 
Number of subgoals & 11 & 11 & 11 & 11 & 3 \\
\bottomrule
\end{tabular}
\end{small}
\end{table}

For LEAP, we re-implemented \cite{nasiriany2019planning} and train TDM \cite{pong2018temporal} policies and Q networks with hidden layers of size $[400, 300]$ and $ReLU$ activation functions.
In the ant navigation environments, we pretrain VAEs with mean squared reconstruction error loss and hidden layers of size $[64, 128, 64]$, $ReLU$ activation functions and representation size of $8$ for the encoders and the decoders.
In the vision based robotic manipulation environment, we pretrain VAEs with mean squared error reconstruction loss and convolutional layers with encoder kernel of sizes $[5, 5, 5]$, encoder strides of sizes $[3,3,3]$, encoder channels of sizes $[16, 16, 32]$, decoder kernel sizes of sizes $[5, 6, 6]$, decoder strides of sizes $[3, 3, 3]$, and decoder channels of sizes $[32, 32, 16]$, representation size of $16$ and $ReLU$ activation functions.
Table \ref{table:LEAPhyperparameters} reports the policy horizon used for each environment as well as the number of subgoals in the test configuration for the results in Figure 6 of the submission. %\ref{fig:antmaze_navigation}.
For the results presented in Figure 7 of the submission, we adapted the number of subgoals according to the difficulty of each configuration. %\ref{fig:HorizonMazes}

%% file: Appendix/AdditionalResults.tex
\section{Additional Results}\label{ap:additionalresults}

\begin{figure}[h]
    \centering
    \includegraphics[width=0.5\linewidth]{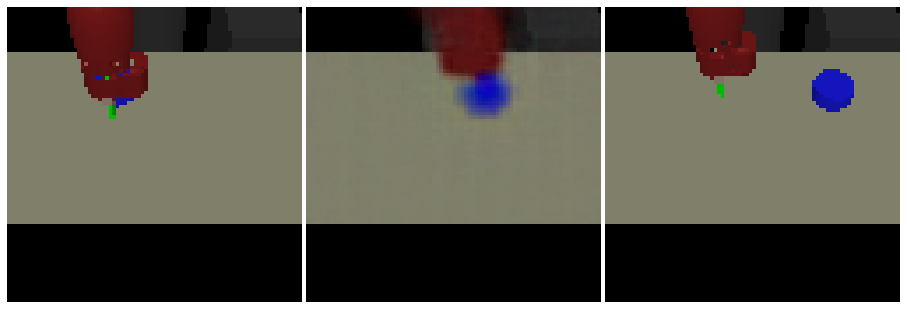}
    
    \centering
    \includegraphics[width=0.5\linewidth]{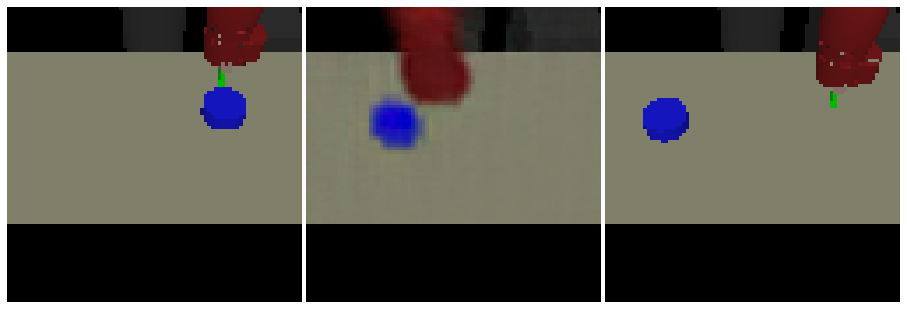}
    
    \centering
    \includegraphics[width=0.5\linewidth]{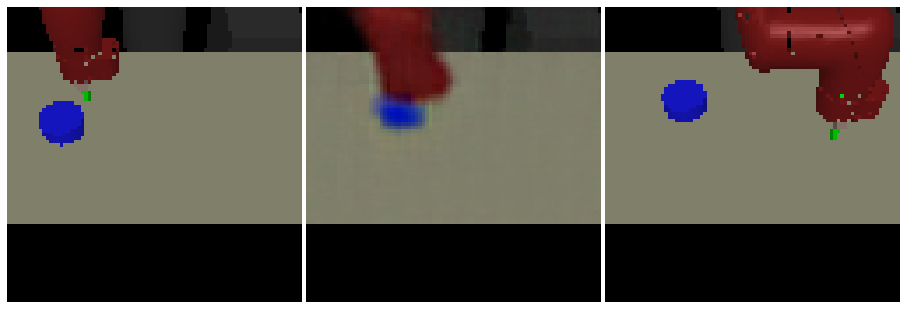}
    
    \centering
    \includegraphics[width=0.5\linewidth]{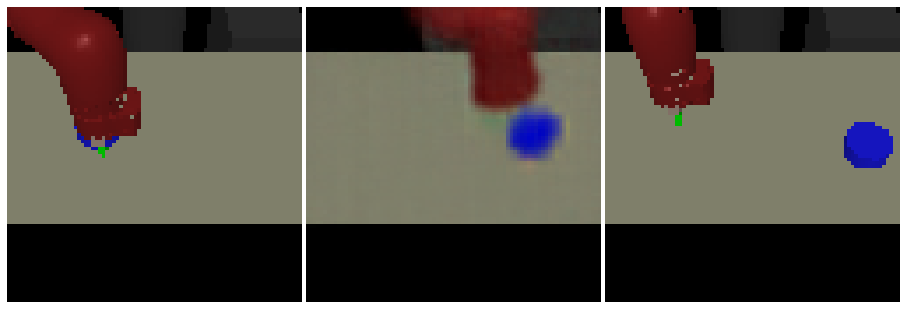}
    
    \centering
    \includegraphics[width=0.5\linewidth]{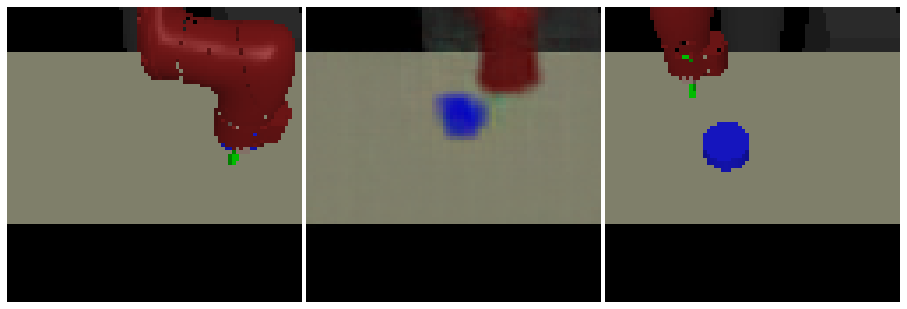}
    
    \caption{Image reconstruction of imagined subgoals (middle column) given current states (left column) and desired goals (right column) for different random configurations in the vision-based robotic manipulation environment.}
    \label{fig:additionalImaginedSubgoals}
\end{figure}

In Figure \ref{fig:additionalImaginedSubgoals}, we provide image reconstructions of imagined subgoals on additional configurations of the vision-based robotic manipulation environment.
